\documentclass{article}

\usepackage{microtype}
\usepackage{graphicx}
\usepackage{subcaption}
\usepackage{booktabs} 

\usepackage{hyperref}

\usepackage[accepted]{icml2026}

\usepackage{amsmath}
\usepackage{amssymb}
\usepackage{mathtools}
\usepackage{amsthm}

\usepackage[capitalize,noabbrev]{cleveref}
\usepackage{xspace}
\theoremstyle{plain}

\theoremstyle{definition}

\theoremstyle{remark}

\usepackage[textsize=tiny]{todonotes}

\usepackage{soul} 

\usepackage[final,commandnameprefix=always]{changes}

\usepackage{multirow}
\usepackage{subcaption}
\usepackage{diagbox}
\usepackage{listings}
\usepackage[most]{tcolorbox} 
\usepackage{xspace}
\tcbuselibrary{listings, skins, breakable}

\definecolor{promptHeader}{RGB}{189, 189, 189} 
\definecolor{promptBody}{RGB}{245, 245, 245}   
\definecolor{promptBorder}{RGB}{0, 0, 0} 

\newtcolorbox{promptbox_base}[1][]{
    colback=gray!5,
    colframe=gray!75,
    colupper=black,
    breakable,
    nobeforeafter,
    enhanced,
    sharp corners,
    boxrule=0.5pt,
    before upper={%
        \fontencoding{T1}\selectfont 
        \catcode`\_=12 \catcode`\$=12 \catcode`\%=12 
        \catcode`\&=12 \catcode`\#=12 \catcode`\^=12
        \catcode`\<=12 \catcode`\>=12
    },
    #1
}

\newenvironment{promptbox}[1]{%
    \begin{promptbox_base}
    {\colorbox{blue!20}{\bfseries \strut #1}}\medskip\par
}{%
    \end{promptbox_base}
}

\newcommand{\modelname}{C-MOP}
\newcommand{\modelfullname}{Cluster-based Momentum Optimized Prompting}
\newcommand{\modelbffullname}{\textbf{C}luster-based \textbf{M}omentum \textbf{O}ptimized \textbf{P}rompting}

\newcommand{\methodone}{Boundary-Aware Contrastive Sampling\xspace}
\newcommand{\methodoneshort}{BACS\xspace}
\newcommand{\methodtwo}{Momentum-Guided Semantic Clustering\xspace}
\newcommand{\methodtwoshort}{MGSC\xspace}

\renewcommand{\algorithmiccomment}[1]{// #1}

\usepackage{tabularx}

\usepackage[most]{tcolorbox}
\usepackage{subcaption}
\usepackage{bbding}
\newtcolorbox{promptcard}[1]{
    colback=promptBody,      
    colframe=promptBorder,    
    colbacktitle=promptHeader, 
    coltitle=black,           
    title={#1},               
    arc=2mm,                  
    boxrule=1.0pt,            
    left=4mm, right=4mm,      
    top=4mm, bottom=4mm,      
    toptitle=1.5mm,           
    bottomtitle=1.5mm,        
    enhanced,                 
    drop shadow={black!20!white}, 
}

\usepackage{hyperref}
\icmltitlerunning{\modelname{}: Integrating Momentum and Boundary-Aware Clustering for Enhanced Prompt Evolution}

\begin{document}


\twocolumn[
  \icmltitle{\modelname{}: Integrating Momentum and Boundary-Aware Clustering\\ for Enhanced Prompt Evolution}



  \icmlsetsymbol{equal}{*}

  \begin{icmlauthorlist}

    \icmlauthor{Binwei Yan}{noahlab}
    \icmlauthor{Yifei Fu}{noahlab}
    \icmlauthor{Mingjian Zhu}{noahlab}
    \icmlauthor{Hanting Chen}{noahlab}
    \icmlauthor{Mingxuan Yuan}{noahlab}
    \icmlauthor{Yunhe Wang \textsuperscript{\Envelope}}{noahlab}
    \icmlauthor{Hailin Hu \textsuperscript{\Envelope}}{noahlab}
  \end{icmlauthorlist}

  \icmlaffiliation{noahlab}{Noah's Ark Lab, Huawei, China}


  \icmlcorrespondingauthor{Yunhe Wang}{yunhe.wang@huawei.com}
  \icmlcorrespondingauthor{Hailin Hu}{hailin.hu@huawei.com}

  \icmlkeywords{Prompt Optimization, Clustering}

  \vskip 0.3in
]



\printAffiliationsAndNotice{}  

\begin{abstract}
Automatic prompt optimization is a promising direction to boost the performance of Large Language Models (LLMs). However, existing methods often suffer from noisy and conflicting update signals. In this research, we propose \textbf{\modelname{}} (\modelbffullname{}), a framework that stabilizes optimization via \methodone{} (\methodoneshort{}) and \methodtwo{} (\methodtwoshort{}). Specifically, \methodoneshort{} utilizes batch-level information to mine tripartite features—\textit{Hard Negatives, Anchors, and Boundary Pairs}—to precisely characterize the typical representation and decision boundaries of positive and negative prompt samples. To resolve semantic conflicts, \methodtwoshort{} introduces a textual momentum mechanism with temporal decay that distills persistent consensus from fluctuating gradients across iterations. Extensive experiments demonstrate that \modelname{} consistently outperforms SOTA baselines like PromptWizard and ProTeGi, yielding average gains of 1.58\% and 3.35\%. Notably, \modelname{} enables a general LLM with 3B activated parameters to surpass a 70B domain-specific dense LLM, highlighting its effectiveness in driving precise prompt evolution. The code is available at \url{https://github.com/huawei-noah/noah-research/tree/master/C-MOP}.

\end{abstract}

\section{Introduction}

Large Language Models (LLMs)~\cite{brown2020language, touvron2023llama, chowdhery2023palm} have become one of the most influential research frontiers in artificial intelligence, demonstrating remarkable capabilities across diverse tasks~\cite{zhao2023survey, minaee2024large}. However, their performance remains highly sensitive to prompt design—a phenomenon that presents both opportunities and challenges for practical deployment~\cite{zhao2021calibrate, lu2022fantastically}. Consequently, an effective prompt optimization strategy is required to help people generate the best possible prompts.


\begin{figure}[!t]
    \begin{subfigure}{1.0\linewidth}
        \includegraphics[width=1.0\linewidth]{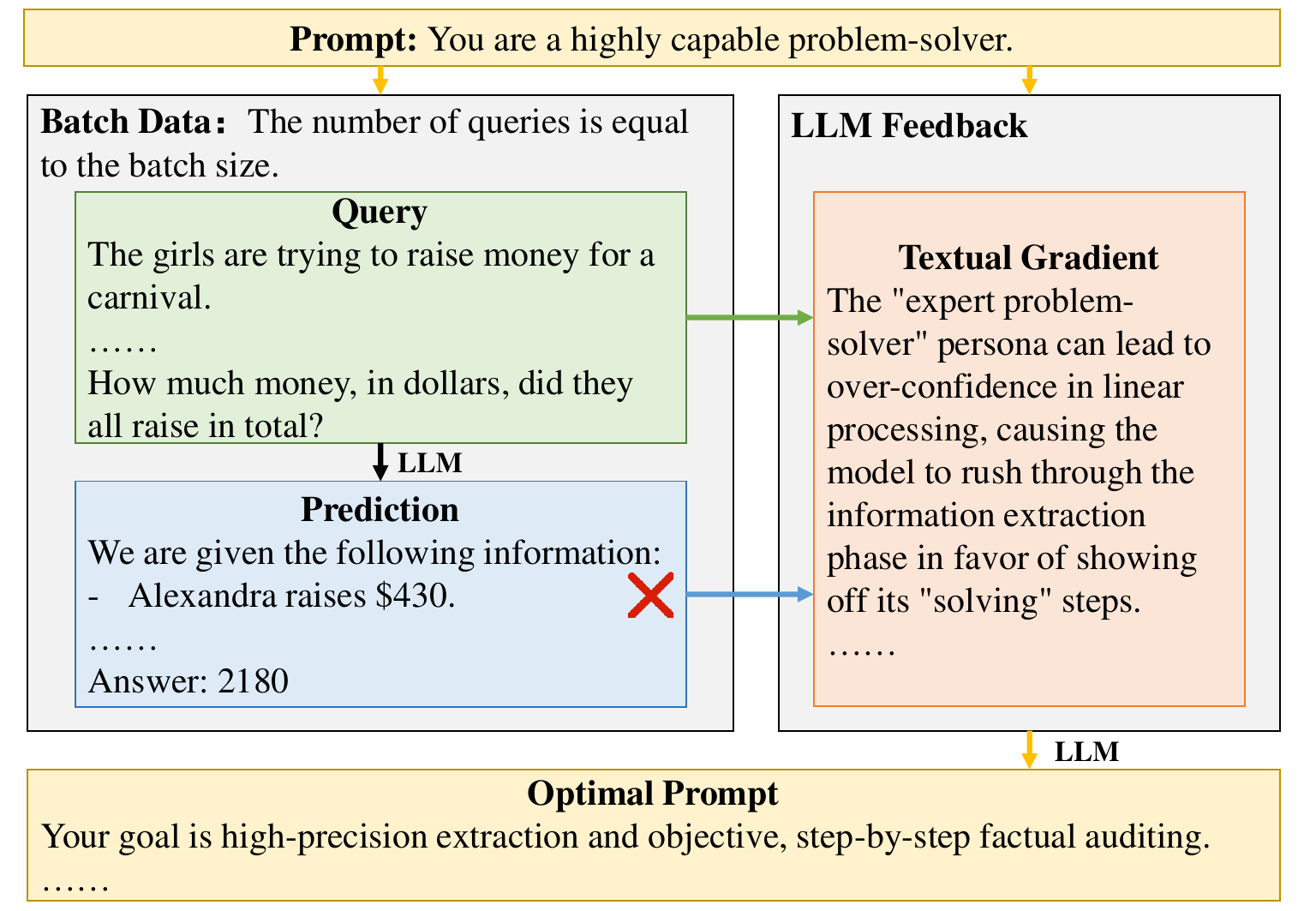}
        \caption{}
        \label{fig:prompt_optimization_workflow}
    \end{subfigure}
    \begin{subfigure}{1.0\linewidth}
        \includegraphics[width=1.0\linewidth]{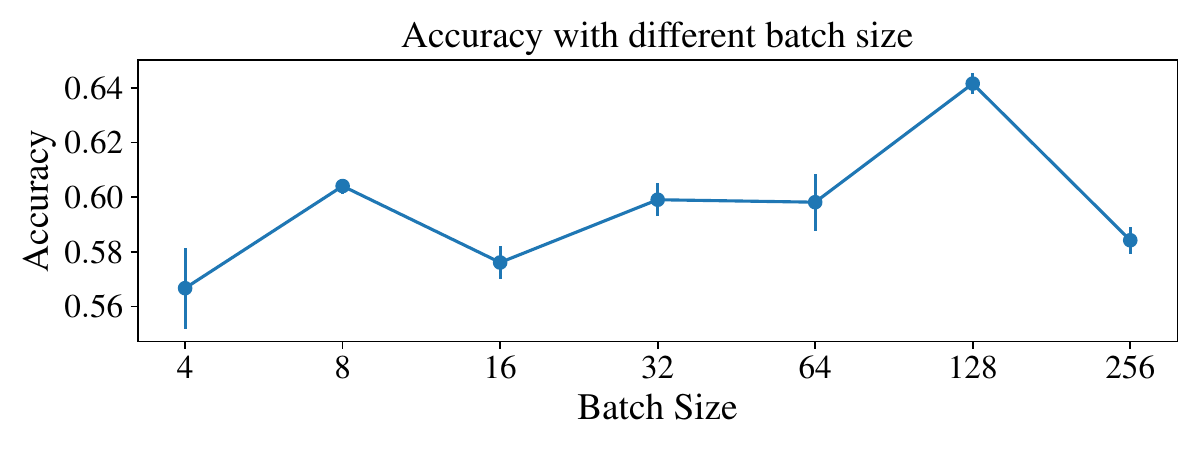}
        \caption{}
    \label{fig:accuracy_for_batchsize}
    \end{subfigure}
    \caption{(a) Textual gradient-based prompt optimization workflow, showing the definition of ``textual gradient'', ``batch size'' and other components. (b) Accuracy versus different batch sizes using Qwen3-30B-A3B-Instruct-2507~\cite{yang2025qwen3} to optimize prompts on the Liar dataset~\cite{wang2017liar}.}
\end{figure}

Early prompt engineering~\cite{jiang2020can, reynolds2021prompt, liu2023pre} relied heavily on manual expertise and trial-and-error, whereas recent research has proposed automatic prompt optimization~\cite{tong2025evoprompt, yu2025premise} to reduce manual burden. Among various approaches, textual gradient-based methods~\cite{pryzant2023automatic, agarwal2025promptwizard} (workflow is illustrated in Figure~\ref{fig:prompt_optimization_workflow}) have gained significant attention due to their ability to provide interpretable directional guidance for iterative refinement. However, the dissection of this ``LLM-as-optimizer'' process~\cite{yang2023large} is generally lagging behind, which hinders this application in real-world applications. 


In back-propagation-based model optimization, the scale of batching is of great importance and generally increasing the batch size can reduce gradient noise and improve training efficiency~\cite{krizhevsky2014one, hoffer2017train, mccandlish2018empirical, shallue2019measuring}. While transferring this idea to textual gradient is intuitively straightforward, till now few studies have explicitly explored this direction. To fill this gap, we first characterize the performance of textual gradient optimization with respect to the inputting example numbers, as illustrated in Figure~\ref{fig:accuracy_for_batchsize}. Intriguingly, the results demonstrate a clear performance gain over 8\% when scaling up the batch size from 4 to 128, and then the size effect saturates to a point that further expansion fails to benefit the performance.

These interesting phenomenon indicate the potential of applying tradition optimization principle to automatic prompt optimization. While enlarging the batch size is generally beneficial, we hypothesize that too many incorrect or non-representative samples will result in gradients that are noisy, unfocused, or even contradictory. Such gradient conflicts hinder optimization stability, making it difficult for the update process to converge to high-quality prompts~\cite{sener2018multi}. Moreover, unlike the numerical averaging operation in gradients of back propagation, expanding the batch size of textual gradients requires populating the context with different exemplars, which is constrained by the context length of LLMs or the memory of hardware. Drawing from these two insights, in this work, we aim to further enhance the effectiveness of large-batch textual gradient optimization with strategic sample selection~\cite{bengio2009curriculum, settles2009active}. In particular, we propose a novel framework method, named \modelname{} (\modelfullname{}), which is designed to identify and select more representative samples. This approach effectively emulates the benefits of an expanded batch size, thereby maximizing the positive impact of batch size increases. Our key innovations are as follows:

\begin{enumerate}
    \item \textbf{\methodone{} (\methodoneshort{})}: 
    We replace stochastic sampling with error-rate-aware clustering to systematically select \textit{Hard Negatives}, \textit{Anchors}, and \textit{Boundary Pairs}. These points collectively map the semantic transition between successful and erroneous regions, thereby delineating the decision boundaries of prompt. By isolating the minimal linguistic differences that trigger classification flips within these pairs, \methodoneshort{} extracts precise contrastive signals that pinpoint specific prompt deficiencies.
    
    \item \textbf{\methodtwo{} (\methodtwoshort{})}: 
    To resolve textual gradient conflicts, we introduce a refinement mechanism that maintains a historical gradient pool with a temporal decay factor. By accumulating gradients across iterations, \methodtwoshort{} ensures the temporal persistence of optimization signals, allowing weighted clustering to extract the semantic consensus while marginalizing batch-specific contradictions. This process transforms transient, fluctuating feedback into a unified and stable optimization direction.
\end{enumerate}

Our extensive experiments on four diverse benchmarks, i.e BBH~\cite{suzgun2022challenging}, GSM8K~\cite{cobbe2021training}, CFinBench~\cite{nie2025cfinbench}, and Liar~\cite{wang2017liar} demonstrate that \modelname{} consistently outperforms state-of-the-art baselines, achieving 1.58\% to 3.35\% average improvement. Ablation studies confirm that clustering contributes substantially to the effectiveness of the framework.

\section{Related Work}

\subsection{Manual Prompt Engineering}
Prompt engineering has long been a central technique for eliciting desired behaviors from large language models. Early work~\cite{reynolds2021prompt, wei2022chain} demonstrated that constructing prompts through methods such as anthropomorphism, analogy, and demonstration to enable capabilities like in-context learning and few-shot reasoning. However, manual prompt design relies heavily on extensive expertise and domain knowledge~\cite{reynolds2021prompt, liu2023pre}, and even subtle linguistic variations can significantly impact model performance.~\cite{brown2020language, jiang2020can, perez2021true}. Consequently, handcrafted prompting is time-consuming and labor-intensive, often requiring exhaustive trial-and-error~\cite{zamfirescu2023johnny}. These drawbacks have motivated a growing interest in automating the prompt creation and optimization process.

\subsection{Automatic Prompt Optimization}
To overcome the limitations of manual prompting, a growing body of work has explored automatic prompt optimization. Early studies proposed trajectory-based optimization~\cite{yang2023large, tang2025unleashing}, which generates new prompts based on historical candidates and their corresponding scores. Evolutionary-based methods~\cite{fernando2023promptbreeder, tong2025evoprompt}, on the other hand, iteratively refine prompts by combining LLMs with evolutionary algorithms. Recent research has shifted its focus toward feedback-based approaches~\cite{pryzant2023automatic, agarwal2025promptwizard}, which extract textual gradients from error analysis as feedback signals to guide prompt updates. Furthermore, to enhance optimization effectiveness, some researchers have adopted multi-agent collaboration to assign diverse roles to LLMs~\cite{han2025mapgd, zhang2025mars, pei2025scope}, while others have applied model optimization techniques to prompt tuning, integrating historical textual gradients via momentum to mitigate noise~\cite{peng2025dlpo}. Nevertheless, they fail to account for the disparity between textual and numerical gradients. Due to their discreteness, textual gradients are more susceptible to mutual interference, thereby undermining the overall optimization effectiveness. Consequently, we aim to employ a more robust update signal selection scheme to ensure optimization stability.

\subsection{Limitations of Existing Methods} Despite these advances, existing textual gradient-based prompt optimization methods face two critical challenges: \begin{enumerate} \item \textbf{Incomplete Boundary Characterization}: Current methods typically rely on sparse, stochastic sampling of a small subset of failure cases~\cite{pryzant2023automatic}. This myopic approach fails to capture the global distribution of model performance and neglects the subtle ``decision boundaries'' where the model fluctuates between success and failure, resulting in unfocused and imprecise gradient signals. \item \textbf{Textual Gradient Conflict and Instability}: In large-scale batch optimization, LLM-generated gradients are often redundant or even contradictory, where different error cases suggest opposing optimization directions~\cite{han2025mapgd}. Without a mechanism to aggregate information across iterations, these methods suffer from significant gradient noise and unstable optimization trajectories. \end{enumerate} To address these limitations, we propose \modelname{}, which introduces a \methodone{} strategy to map precise decision boundaries and a textual momentum mechanism that leverages historical gradients to stabilize the optimization process.

\section{Method}
In this section, we present our proposed framework for automated prompt optimization. 

\begin{algorithm}[ht]
    \small
    \caption{\modelname{}: \modelfullname{}}
    \label{alg:cmop}
    \begin{algorithmic}[1]
        \STATE {\bfseries Input:} Task model $\mathcal{F}$, Optimizer model $\mathcal{F}_{opt}$, Dataset $\mathcal{D}$, Initial prompt $p_0$, Batch size $B$, Iterations $T$, Total quota $Q_{total}$, Decay factor $\gamma$, Embedding model $\phi$.
        \STATE {\bfseries Output:} Optimized prompt $p_T$.
        \STATE Initialize Gradient Pool $\mathcal{G}_{pool} \leftarrow \emptyset$ 
        
        \FOR{$t = 0$ {\bfseries to} $T-1$}
            \STATE Sample batch $\mathcal{D}_t \subset \mathcal{D}$ with size $B$.
            \STATE Generate predictions $\{r_i\}$ using $p_t$ for all $(q_i, a_i) \in \mathcal{D}_t$.
            
            \STATE Compute embeddings $e_i = \phi(\text{concat}(q_i, a_i, r_i))$ and cluster into $\{\mathcal{C}_1, \ldots, \mathcal{C}_K\}$.
            \FOR{each cluster $\mathcal{C}_k$}
                \STATE Calculate error rate $E_k$ and assign sample quota $Q_k$ for cluster $\mathcal{C}_k$.
                \STATE $\mathcal{S}_k \leftarrow$ Sample $Q_k$ instances focusing on the tripartite structures.
                \STATE $\mathcal{G}_{new} \leftarrow \text{GenerateGradients}(\mathcal{F}_{opt}, p_t, \mathcal{S}_k)$.
            \ENDFOR
            
            \STATE $\mathcal{G}_{pool} \leftarrow \mathcal{G}_{pool} \times \gamma$ 
            \STATE Add $\mathcal{G}_{new}$ to $\mathcal{G}_{pool}$ with initial weights.
            \STATE Cluster $\mathcal{G}_{pool}$ and aggregate weights within each cluster.
            \STATE $\mathcal{T}_t \leftarrow$ Select top $n$ gradients with highest weights.
            
            \STATE $\mathcal{P}_{cand} \leftarrow \emptyset$

            \FOR{$l = 1$ {\bfseries to} $L$}
                \STATE Select $g^* \in \mathcal{T}_t$ and generate $\hat{p}_l$ via $\mathcal{F}_{opt}$ rewriting $p_t$.
                \STATE $\mathcal{P}_{cand} \leftarrow \mathcal{P}_{cand} \cup \{\hat{p}_l\}$
            \ENDFOR
            \STATE $p_{t+1} \leftarrow \text{UCB\_Select}(\mathcal{P}_{cand})$
        \ENDFOR
        \STATE {\bfseries return} $p_T$
    \end{algorithmic}
\end{algorithm}

\begin{figure*}[htbp]
    \centering
    \includegraphics[width=1.0\textwidth]{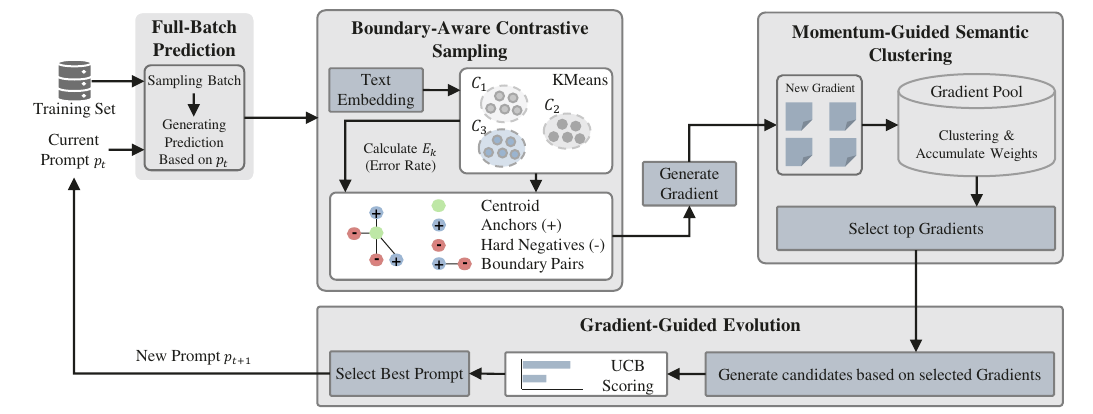}
    \caption{Overview of the \modelname{} framework. The pipeline consists of four stages: (1) Full-Batch Prediction; (2) \methodone{}; (3) \methodtwo{}; (4) Gradient-Guided Evolution.}
    \label{fig:arch}
\end{figure*}





\subsection{Problem Formulation}

Let $\mathcal{D} = \{(q_i, a_i)\}_{i=1}^N$ denote a dataset of query-answer pairs, where $q_i$ represents an input query and $a_i$ the corresponding ground-truth answer. Given a base language model $\mathcal{F}$ and an initial prompt $p_0$, our goal is to find an optimized prompt $p^*$ that maximizes the expected task performance:

\begin{equation}
p^* = \arg\max_{p} \mathbb{E}_{(q,a) \sim \mathcal{D}} [s(\mathcal{F}(p, q), a)]
\end{equation}

where the expectation $\mathbb{E}_{(q,a) \sim \mathcal{D}}$ represents the average performance over the data distribution. In practice, this is approximated by evaluating the scoring function $s(\cdot, \cdot)$ (e.g., accuracy, F1 score) across sampled batches of instances. The challenge lies in navigating the discrete, high-dimensional prompt space to find a prompt that generalizes across the entire distribution while avoiding local optima~\cite{bengio2013estimating}.

\subsection{Framework Overview}

\modelname{} operates through an iterative refinement process consisting of four key stages, as illustrated in Figure~\ref{fig:arch}:

\begin{enumerate}

    \item \textbf{Full-Batch Prediction}: Evaluate the current prompt $p_t$ on an entire sampled batch. Unlike traditional methods that only store failure cases, we preserve both positive and negative samples to provide a global view of the performance of the current prompt.
    

    \item \textbf{\methodone{} (\methodoneshort{})}: We apply semantic clustering to the batch and allocate sampling quotas based on cluster-level error rates. Within each cluster, we extract \textit{Hard Negatives}, \textit{Anchors}, and \textit{Boundary Pairs} to capture the subtle decision logic of the model.
    

    \item \textbf{\methodtwo{} (\methodtwoshort{})}: We generate textual gradients~\cite{pryzant2023automatic} for the sampled cases. To resolve gradient conflicts, we maintain a historical gradient pool with a temporal decay $\gamma$, performing secondary clustering to select the most consistent and effective optimization directions.

    
    \item \textbf{Gradient-Guided Evolution}: Guided by the refined gradients, the optimizer model generates a diverse set of candidate prompts $\mathcal{P}_{cand}$. To efficiently navigate this pool, we employ a UCB-based strategy~\cite{auer2002finite} that balances exploration and exploitation, selecting the top-$W$ prompts for the next iteration.
    
\end{enumerate}


We now detail each component, with particular emphasis on our \methodone{} and \methodtwo{} evolution mechanisms. The complete optimization workflow of the \modelname{} framework is formally presented in Algorithm~\ref{alg:cmop}.



\subsection{\methodone{}}


To ensure high-fidelity feedback, \methodoneshort{} replaces stochastic sampling with a systematic extraction of \textit{Hard Negatives}, \textit{Anchors}, and \textit{Boundary Pairs}. The motivation for this design is two-fold: first, \textit{Anchors} are essential for establishing a stable reference of successful reasoning, preventing the optimizer model from drifting away from correct logic during refinement; second, \textit{Boundary Pairs} are crucial for pinpointing the precise linguistic triggers that cause the model to flip between correct and incorrect outputs. By contrasting these prototypical success patterns with subtle failure modes, \methodoneshort{} precisely characterizes the decision boundaries of prompt to generate high-contrast, boundary-aware textual gradients. The strategy follows four sequential steps to select the most informative samples from a batch $\mathcal{B}$:

1. \textbf{Global Batch Embedding}: For every sample $(q_i, a_i)$ in the batch, where $r_i$ is the prediction of task model, we generate a semantic representation $e_i$ using a pre-trained encoder $\phi$: \begin{equation} e_i = \phi(\text{concat}(q_i, a_i, r_i)) \end{equation} By embedding the entire batch, we create a complete semantic map of the prompt’s current behavior across different input types.

2. \textbf{Cluster-Aware Quota Allocation}: We apply K-means clustering~\cite{krishna1999genetic} to partition the batch into $K$ clusters $\{C_1, C_2, \dots, C_K\}$ and sample a total of $Q_{total}$ instances from a single batch. For each cluster $C_k$, we calculate its error rate $E_k = \frac{|\mathcal{B}_{error} \cap C_k|}{|C_k|}$. We then allocate a sampling quota $Q_k$ proportional to the error rate: \begin{equation} Q_k = \max \left( 1, \left\lceil Q_{total} \times \frac{E_k}{\sum_{j=1}^K E_j} \right\rceil \right) \end{equation} This mechanism ensures that clusters exhibiting higher failure rates—which highlight the logical deficiencies of the current prompt—are prioritized for further optimization.

    
    

3. \textbf{Tripartite Fine-Grained Sampling}: Within each cluster $C_k$, we define positive samples $C_k^+$ as correctly predicted instances and negative samples $C_k^-$ as incorrectly predicted ones. Based on the quota $Q_k$, we select three types of samples to provide the optimizer model with maximal contrast:
\begin{itemize}
    \item \textbf{Hard Negatives}: The negative sample closest to the centroid. Since the centroid represents the dominant semantic theme of the cluster, these samples identify systematic failures rather than random outliers, representing cases where the reasoning logic of the prompt consistently fails:
    \begin{equation}
        e^{hard} = \arg\min_{e \in C_k^-} \|e - \mu_k\|^2
    \end{equation}
    \item \textbf{Anchors}: The positive sample closest to the centroid $\mu_k$. These instances represent the prototypical success patterns of the cluster, establishing a stable baseline for the correct reasoning logic of the prompt:\begin{equation}e^{anc} = \arg\min_{e \in C_k^+} \|e - \mu_k\|^2\end{equation}
    \item \textbf{Boundary Pairs}: The $n$ pairs of samples $(e^+, e^-)$ that minimize the semantic distance $\|e^+ - e^-\|^2$. These pairs are critical for decision boundary refinement; by isolating the minimal semantic difference that triggers a classification flip, they provide high-contrast signals that pinpoint the exact instructional flaws.
\end{itemize}

4. \textbf{Gradient Generation}: The selected samples are passed to the optimizer model. By comparing the \textit{Hard Negatives} and \textit{Anchors} alongside the \textit{Boundary Pairs}, the LLM can derive highly actionable gradients that specifically target the confusion between similar semantic inputs. 

\subsection{\methodtwo{}}

We propose \methodtwo{} (\methodtwoshort{}), a mechanism that maintains a historical gradient pool through a temporal decay factor. By aggregating gradients across iterations, \methodtwoshort{} distills persistent semantic consensus into a unified and stable direction for prompt evolution. The refinement process is formalized as follows:

1. \textbf{Dynamic Gradient Pool Maintenance}: We maintain a historical gradient pool $\mathcal{G}_t = \{ (g_i, w_i, t_i) \}$, where each entry consists of a textual gradient $g_i$, its importance weight $w_i$, and the iteration index $t_i$ when the gradient is generated. At each iteration $t$, all historical gradients undergo a temporal decay:
\begin{equation}
w_i^{(t)} = w_i^{(0)} \cdot \gamma^{(t - t_i)}
\end{equation}
where $\gamma$ is a decay factor. This ensures that more recent gradients—which reflect the model's current behavior—carry stronger influence, while older gradients are gradually phased out.

2. \textbf{Cross-Iteration Gradient Clustering}: 
To resolve textual conflicts and ensure temporal consistency, we perform semantic clustering on the unified set of current gradients and the decayed historical pool. The total gradient set is defined as:
\begin{equation}
\mathbf{G}_{total} = \mathcal{G}{current} \cup \mathcal{G}{pool}
\end{equation}
We subsequently partition $\mathbf{G}_{total}$ into $M$ semantic clusters $\{\hat{C}_1, \hat{C}_2, \dots, \hat{C}_M\}$ based on their latent embeddings. This clustering allows the framework to identify persistent optimization themes across different iterations.

3. \textbf{Weight-Based Gradient Selection}: Within each cluster $\hat{C}_j$, gradients may still point in slightly different directions. To extract the most representative and ``stable'' direction, we select the top-$n$ gradients with the highest accumulated weights:
\begin{equation}
g^*_j = \text{Top-n}_{g \in \hat{C}_j} (w^{(t)})
\end{equation}
This selection process acts as a semantic moving average, effectively suppressing stochastic noise from individual batches.

4. \textbf{Prompt Evolution}: The final set of refined gradients $\{g^*_j\}$ is used to guide the optimizer model. By focusing on high-weight, cluster-representative gradients, the evolution process remains coherent across iterations, preventing the ``oscillation'' common in naive textual gradient descent.

\begin{table*}[htbp]
    \centering
    \caption{Main Results: Accuracy comparison (\%) on three benchmarks. Best results are \textbf{bolded}, second-best are \underline{underlined}. Results are averaged over three runs with standard deviation shown as subscripts. Asterisks (*) denote our re-implemented version. Q30BA3B-Inst: Qwen3-30B-A3B-Instruct-2507, Q30BA3B-Think: Qwen3-30B-A3B-Thinking-2507.}
    \label{tab:main_results}
    \small
    \renewcommand{\arraystretch}{1.09}
    \begin{tabular*}{\textwidth}{@{\extracolsep{\fill}} lcccccc}
        \toprule
        
        \multirow{2}{*}{\textbf{Method}} & \multirow{2}{*}{\textbf{Task Model}} & \multirow{2}{*}{\textbf{Optimizer Model}} & \textbf{BBH} & \textbf{GSM8K} & \textbf{Liar} & \multirow{2}{*}{\textbf{Avg}} \\
        & & & \textbf{(EM)} & \textbf{(Acc)} & \textbf{(F1)} & \\
        \midrule
        OPRO~\cite{yang2023large} & PaLM 2-L & PaLM 2-L-IT & - & 80.20 & - & - \\
        EvoPrompt~\cite{tong2025evoprompt} & GPT-3.5-Turbo & GPT-3.5-Turbo & 75.03 & - & - & - \\
        PromptWizard~\cite{agarwal2025promptwizard} & GPT-4 & GPT-4 & 88.10 & 95.40 & - & - \\
        SPO~\cite{xiang2025self} & GPT-4o-mini & GPT-4o & - & - & 66.90 & - \\
        ERM~\cite{yan2025efficient} & Doubao-Pro & GPT-4o & 86.10 & 93.30 & 68.60 & - \\
        MARS~\cite{zhang2025mars} & GPT-4o & GPT-4o & 80.86 & - & - & - \\
        GPO~\cite{tang2025unleashing} & GPT-4 & GPT-4 & 78.65 & - & - & - \\
        \midrule \midrule
        Zero-shot (Base) & Q30BA3B-Inst & - & 81.65$_{\pm 0.14}$ & 80.67$_{\pm 0.47}$ & 40.92$_{\pm 0.33}$ & 67.75 \\
        ProTeGi*~\cite{pryzant2023automatic} & Q30BA3B-Inst & Q30BA3B-Think & 82.63$_{\pm 1.27}$ & \underline{95.07}$_{\pm 0.20}$ & 58.47$_{\pm 0.43}$ & 78.72 \\
        PromptWizard* & Q30BA3B-Inst & Q30BA3B-Think & \underline{85.34}$_{\pm 1.51}$ & 94.54$_{\pm 0.68}$ & \underline{61.61}$_{\pm 0.62}$ & \underline{80.50} \\
        SPO* & Q30BA3B-Inst & Q30BA3B-Think & 84.13$_{\pm 1.31}$ & 93.00$_{\pm 1.01}$ & 59.38$_{\pm 3.07}$ & 78.84 \\
        GPO* & Q30BA3B-Inst & Q30BA3B-Think & 82.61$_{\pm 1.61}$ & 95.04$_{\pm 0.79}$ & 57.85$_{\pm 3.53}$ & 78.50 \\
        \midrule
        \textbf{\modelname{} (Ours)} & Q30BA3B-Inst & Q30BA3B-Think & \textbf{86.24}$_{\pm 0.41}$ & \textbf{95.53}$_{\pm 0.20}$ & \textbf{64.46}$_{\pm 0.68}$ & \textbf{82.08} \\
        $\Delta$ vs. PromptWizard & & & +0.90 & +0.99 & +2.85 & +1.58 \\
        \bottomrule
    \end{tabular*}
\end{table*}

\subsection{Gradient-Guided Evolution}

The final stage of our framework transforms the stable semantic gradients into concrete prompt instructions through a two-fold evolution process: candidate generation and adaptive selection.

\textbf{Candidate Generation.} The diversity and quality of the candidate pool are fundamental to the robustness of the optimization. To ensure a broad yet targeted search within the prompt manifold, we derive $2M$ primary optimization directions by selecting the top-2 gradients with the highest accumulated weights from each of the $M$ semantic clusters in \methodtwoshort{}:
\begin{equation}
    \mathcal{G}_{refined} = \{ g^*_{j,k} \mid j \in \{1, \dots, M\}, k \in \{1, 2\} \}
\end{equation}
These gradients represent stabilized, cluster-level insights that allow the optimizer model to focus on rectifying systemic logical deficiencies rather than isolated failures. For each $g^* \in \mathcal{G}_{refined}$, the optimizer model rewrites the current prompt $p_t$ to produce a total of $L$ distinct candidates $\mathcal{P}_{cand}$. During this process, we maintain an appropriate sampling temperature to encourage structural and vocabulary diversity, ensuring that candidates are not merely minor paraphrases but distinct strategic attempts at task instruction.

\textbf{UCB-Driven Selection.} Given the large-scale candidate prompt pool, evaluating every prompt on the entire validation set would be computationally prohibitive. To efficiently navigate this search space, we frame the selection as a multi-armed bandit problem and employ the Upper Confidence Bound (UCB) algorithm~\cite{auer2002finite}:
\begin{equation}
    UCB(p) = \mu(p) + \alpha \sqrt{\frac{\ln N}{n_p}}
\end{equation}
where $\mu(p)$ is the empirical mean accuracy, $n_p$ is the number of evaluations for prompt $p$, and $N$ is the total evaluation trials. This mechanism dynamically balances the exploitation of high-performing prompts with the exploration of candidates whose potential remains uncertain. By iteratively assessing candidates on fresh subsets of data until the budget is exhausted, \modelname{} prevents the optimization from converging prematurely to local optima. Finally, the top-$W$ prompts with the highest empirical performance are selected as the ``beam'' for the subsequent iteration.

\section{Experiments}
\label{sec:experiments}

In this section, we conduct a comprehensive empirical evaluation to validate the effectiveness of \modelname{}. Our experiments aim to answer four key research questions:

\begin{enumerate}
    \item \textbf{RQ1}: How does \modelname{} compare to state-of-the-art prompt optimization methods?
    \item \textbf{RQ2}: What is the individual contribution of each component? 
    \item \textbf{RQ3}: How does the choice of optimizer model affect optimization outcomes?
    \item \textbf{RQ4}: Can \modelname{} enable general-purpose models to rival domain-specific models?
\end{enumerate}


\subsection{Experimental Setup}

\textbf{Benchmarks.} We evaluate \modelname{} across four diverse datasets to demonstrate its applicability: BBH~\cite{suzgun2022challenging} (Big-Bench Hard) for complex logical reasoning, Liar~\cite{wang2017liar} for short-statement fact-checking, GSM8K~\cite{cobbe2021training} for multi-step mathematical problems, and CFinBench~\cite{nie2025cfinbench} for specialized financial domain knowledge. More detailed descriptions and data statistics for each benchmark are presented in Appendix~\ref{app:datasets_details}.

\textbf{Baselines.} We compare \modelname{} with Zero-shot (with a default prompt) and four re-implemented existing methods, including ProTeGi~\cite{pryzant2023automatic}, PromptWizard~\cite{agarwal2025promptwizard}, SPO~\cite{xiang2025self} and GPO~\cite{tang2025unleashing}.


\textbf{Implementation Details.} Unless otherwise specified, we employ \textbf{Qwen3-30B-A3B-Instruct-2507}~\cite{yang2025qwen3} as the task model, while using \textbf{Qwen3-30B-A3B-Thinking-2507}~\cite{yang2025qwen3} as the optimizer model. For text clustering, \textbf{all-MiniLM-L6-v2}~\cite{reimers2019sentence} is used as the embedding model. A default batch size of $B=128$ is employed throughout our experiments. 
The optimization process is conducted over $T=20$ iterations with a beam size of $W=4$. In each iteration, the optimizer model generates a total of $L=10$ candidate prompts based on the refined gradients. We set the total sampling quota $Q_{total}=30$. The number of clusters for \methodoneshort{} is set to $K=14$, while the number of clusters for \methodtwoshort{} is $M=10$. Furthermore, we incorporate a decay factor $\gamma=0.9$ during the gradient aggregation phase to emphasize recent feedback signals. Detailed sensitivity analyses for these hyper-parameters, along with generalization experiments across diverse task models, are presented in Appendix~\ref{app:implemetation_details}. All experiments were conducted using three random seeds, and we report the mean ± standard deviation.

\subsection{Main Results (RQ1)}

Table~\ref{tab:main_results} presents the performance of \modelname{} compared to state-of-the-art (SOTA) methods and re-implemented baselines across three representative benchmarks. The results demonstrate that \modelname{} consistently achieves the highest accuracy among all models using the same task model. We summarize the key findings as follows:

\begin{itemize}
    \item \textbf{Consistency Across Benchmarks:} \modelname{} outperforms all re-implemented baselines on BBH, GSM8K, and Liar. Compared to the strongest baseline, PromptWizard*, \modelname{} achieves absolute gains of +0.90\%, +0.99\%, and +2.85\% respectively. The substantial improvement on the Liar dataset (from 61.61\% to 64.46\%) is particularly noteworthy, highlighting its effectiveness in handling nuanced semantic classification tasks.
    
    \item \textbf{High-Performance Reasoning:} In logic-intensive tasks like GSM8K and BBH, \modelname{} reaches 95.53\% and 86.24\% accuracy. Remarkably, our method using the Qwen3-30B-A3B-Instruct-2507 surpasses several GPT-4 based SOTA methods, such as MARS (80.86\%) and GPO (78.65\%) on the BBH benchmark. This indicates that our optimization strategy can significantly narrow the gap between open-source and proprietary models.
    
    
    \item \textbf{Comparison with Zero-shot Base:} Compared to the Zero-shot (Base) performance, \modelname{} provides a massive boost, particularly on the Liar dataset, where accuracy increases from 40.92\% to 64.46\% (+23.54\%). This underscores the necessity of systematic prompt optimization in specialized tasks where default instructions often fall short.
\end{itemize}

In summary, the empirical results confirm that \modelname{} is a robust and effective framework for prompt optimization, achieving a new state-of-the-art performance on the Qwen3-30B-A3B-Instruct-2507 as the task model.

\begin{table}[htbp]
    \centering
    \caption{Ablation study of \methodtwoshort{} and different instance components. Clustering: K-means clustering on failure instances.}
    \label{tab:ablation_mgsc}
    \small
    \renewcommand{\arraystretch}{1.1}
    \begin{tabular*}{\columnwidth}{@{\extracolsep{\fill}} ccccc}
        \toprule
        
        \multirow{2}{*}{\textbf{Instance}} & \multirow{2}{*}{\textbf{Gradient}} & \textbf{Liar} & \textbf{BBH} & \multirow{2}{*}{\textbf{Avg}} \\
         &  & \textbf{(F1)} & \textbf{(EM)} & \\

        \midrule
        \multicolumn{2}{c}{Zero-shot (Base)} & 40.92 & 81.65 & 61.29 \\
        \midrule
        - & \methodtwoshort{} & 59.84 & 83.23 & 71.54 \\
        Clustering & \methodtwoshort{} & 61.42 & 82.60 & 72.01 \\
        \methodoneshort{} & \methodtwoshort{} & \textbf{64.46} & \textbf{86.24} & \textbf{75.35} \\
        \bottomrule
    \end{tabular*}
\end{table}


\begin{table}[htbp]
    \centering
    \caption{Ablation study of different sample types. HN, ANC, and BP denote \textit{Hard Negatives}, \textit{Anchors}, and \textit{Boundary Pairs}, respectively.}
    \label{tab:ablation_bacs}
    \small
    \renewcommand{\arraystretch}{1.1}
    \setlength{\tabcolsep}{2pt}
    \begin{tabularx}{\columnwidth}{@{\extracolsep{\fill}} cccccc}
        \toprule

        \multirow{2}{*}{\textbf{HN}} & \multirow{2}{*}{\textbf{ANC}} & \multirow{2}{*}{\textbf{BP}} & \textbf{Liar} & \textbf{BBH} & \multirow{2}{*}{\textbf{Avg}} \\
         &  &  & \textbf{(F1)} & \textbf{(EM)} & \\
        
        \midrule
        \checkmark & & & 61.37 & 84.85 & 73.11 \\
        \checkmark & \checkmark & & 62.39 & 83.65 & 73.02 \\
        \checkmark & & \checkmark & 63.28 & 85.18 & 74.23 \\
        \checkmark & \checkmark & \checkmark & \textbf{64.46} & \textbf{86.24} & \textbf{75.35} \\
        \bottomrule
    \end{tabularx}
\end{table}

\begin{table}[htbp]
    \centering
    \caption{Ablation study on the application of different components to instances and gradients. Momentum: momentum application without semantic gradient clustering~\cite{peng2025dlpo}.}
    \label{tab:ablation}
    \small
    \renewcommand{\arraystretch}{1.1}
    \begin{tabular*}{\columnwidth}{@{\extracolsep{\fill}} ccccc}
        \toprule
        
        \multirow{2}{*}{\textbf{Instance}} & \multirow{2}{*}{\textbf{Gradient}} & \textbf{Liar} & \textbf{BBH} & \multirow{2}{*}{\textbf{Avg}} \\
         &  & \textbf{(F1)} & \textbf{(EM)} & \\

        \midrule
        \multicolumn{2}{c}{Zero-shot (Base)} & 40.92 & 81.65 & 61.29 \\
        \midrule
        Clustering & - & 61.71 & 82.52 & 72.12 \\
        \methodoneshort{} & - & 60.38 & 83.97 & 72.18 \\
        \methodoneshort{} & Momentum & 61.11 & 82.53 & 71.82 \\
        \methodoneshort{} & \methodtwoshort{} & \textbf{64.46} & \textbf{86.24} & \textbf{75.35} \\
        \bottomrule
    \end{tabular*}
\end{table}

\subsection{Ablation Analysis (RQ2)}
\label{sec:ablation}
To verify the individual contributions of our proposed components, we conduct a systematic ablation study on the Liar and BBH datasets (Table~\ref{tab:ablation_mgsc}, ~\ref{tab:ablation_bacs} and~\ref{tab:ablation}). We analyze the impact of different module configurations as follows:
\begin{itemize}
    \item \textbf{Effectiveness of \methodoneshort{}:} 
    As shown in Table~\ref{tab:ablation_mgsc}, keeping the \methodtwoshort{} module constant while replacing \methodoneshort{} with random sampling leads to a drop in average accuracy to 71.54\%, which is significantly lower than the 75.35\% achieved by our full framework. Similarly, employing naive error clustering instead of \methodoneshort{} only yields a 72.01\% average accuracy. This performance gap demonstrates that mining specific \textit{Boundary Pairs} provides more precise directional guidance for prompt evolution than indiscriminately using failure cases.


    \item \textbf{Decomposition of Sample Types in \methodoneshort{}:}
    Table~\ref{tab:ablation_bacs} reveals distinct optimization dynamics between single-domain and multi-task scenarios. On the Liar dataset, \textit{Anchors} and \textit{Boundary Pairs} provide straightforward cumulative gains. However, on the BBH benchmark, 
    adding \textit{Anchors} alone leads to a performance dip from 84.85\% to 83.65\%. We hypothesize that in multi-task settings, a limited set of \textit{Anchors} may overfit to specific sub-tasks, introducing task-specific bias that harms global generalization. Crucially, the integration of \textit{Boundary Pairs} rectifies this instability, propelling the performance to a peak of 86.24\%. This confirms that while \textit{Anchors} offer static guidance, \textit{Boundary Pairs} provide task-agnostic contrastive signals---learning the fundamental distinction between correct and incorrect reasoning---which is essential for robust optimization across diverse tasks.

    \item \textbf{Impact of \methodtwoshort{} Gradient Refinement:} 
    Table~\ref{tab:ablation} highlights the necessity of our gradient refinement mechanism. The complete removal of \methodtwoshort{} results in a reduced average accuracy of 72.18\%. Furthermore, we observe that applying momentum~\cite{peng2025dlpo} without secondary semantic clustering leads to a performance dip on BBH to 82.53\%, an outcome even lower than the 83.97\% attained by \methodoneshort{} alone. This confirms that while momentum stabilizes the trajectory, the secondary clustering in \methodtwoshort{} is essential to filter out batch-specific noise and extract a unified semantic consensus, ultimately driving the average accuracy to its peak of 75.35\%.
\end{itemize}

Overall, these results validate that \methodoneshort{} and \methodtwoshort{} are highly complementary, with the former providing high-quality raw signals and the latter ensuring their stable refinement into consistent optimization directions.

\begin{table}[htbp]
    \centering
    \caption{Impact of optimizer model choice on optimization effectiveness (Liar and BBH dataset). Q30BA3B-Inst: Qwen3-30B-A3B-Instruct-2507, Q30BA3B-Think: Qwen3-30B-A3B-Thinking-2507.}
    \label{tab:llm_impact}
    \small
    \renewcommand{\arraystretch}{1.1}
    \setlength{\tabcolsep}{2pt}
    \begin{tabular*}{\columnwidth}{@{\extracolsep{\fill}} lccc}
        \toprule
        
        \multirow{2}{*}{\textbf{Task Model}} & \multirow{2}{*}{\textbf{Optimizer Model}} & \textbf{Liar} & \textbf{BBH} \\
        & & \textbf{(F1)} & \textbf{(EM)} \\
        \midrule
        Q30BA3B-Inst & N/A (Zero-shot) & 40.92 & 81.65 \\
        Q30BA3B-Inst & Q30BA3B-Inst (Self) & 58.24 & 83.52 \\
        Q30BA3B-Inst & Q30BA3B-Think & \textbf{64.46} & \textbf{86.24} \\
        \bottomrule
    \end{tabular*}
\end{table}

\subsection{Impact of Optimizer Capacity (RQ3)}
\label{subsec:optimizer_impact}

We investigate how the reasoning capacity of the optimizer model affects the effectiveness of prompt evolution. Table~\ref{tab:llm_impact} compares the performance when using different models as the optimizer while maintaining Qwen3-30B-A3B-Instruct-2507 as the task model. Our observations are as follows:

\begin{itemize}
    \item \textbf{Effectiveness of Self-Correction:} Using the task model itself as the optimizer model yields substantial gains over the zero-shot baseline, with accuracy jumping from 40.92\% to 58.24\% on Liar (+17.32\%) and from 81.65\% to 83.52\% on BBH (+1.87\%). This confirms that \modelname{} enables models to effectively identify and rectify their own systemic biases through our cluster-aware gradient mechanism, providing a cost-effective solution without relying on external teacher models.
    
    \item \textbf{Benefits of Enhanced Reasoning:} Upgrading to the Thinking variant further boosts performance to 64.46\% on Liar and 86.24\% on BBH. This significant delta suggests that the quality of textual gradients is highly dependent on the optimizer's ability to perform deep logical deduction. Stronger models generate more precise and actionable gradients, which in turn guide the evolution toward more robust prompt candidates.
    
\end{itemize}

These results validate that \modelname{} is scalable and can effectively utilize stronger optimizer models to guide optimization of weaker task models.

\begin{table}[htbp]
    \centering
    \caption{Comparison with domain-specific models on CFinBench. Results for XuanYuan2-70-Base and GPT-4 are retrieved from~\cite{nie2025cfinbench}. Q30BA3B-Inst: Qwen3-30B-A3B-Instruct-2507. Opt: Optimization.}
    \label{tab:vertical}
    \small
    \renewcommand{\arraystretch}{1.1}
    \begin{tabular*}{\columnwidth}{@{\extracolsep{\fill}} lcc}
        \toprule
        \textbf{Model} & \textbf{Type} & \textbf{Accuracy (\%)} \\
        \midrule
        XuanYuan2-70B-Base & Domain-Specific & 56.69 \\
        GPT-4 & Closed-Source & 56.77 \\
        \midrule
        Q30BA3B-Inst & General & 52.49 \\
        \textbf{+ \modelname{}} & \textbf{General + Opt} & \textbf{60.20} \\
        \bottomrule
    \end{tabular*}
\end{table}

\subsection{Domain Adaptation Comparison (RQ4)}

We examine whether \modelname{} can empower general-purpose models to rival or even surpass domain-specific models. Table~\ref{tab:vertical} compares \modelname{}-optimized Qwen3-30B-A3B-Instruct-2507 against specialized financial models and closed-source SOTA on the CFinBench benchmark. \modelname{} boosts the performance of the general-purpose Qwen3-30B-A3B-Instruct-2507 from 52.49\% to 60.20\%, an absolute gain of +7.71\%. Remarkably, our optimized model with only 3B active parameters surpasses XuanYuan2-70B-Base~\cite{zhang2023xuanyuan} by 3.51\%, despite the latter being a large-scale model specifically pre-trained on expansive financial corpora. It also surpasses GPT-4~\cite{achiam2023gpt} (+3.43\%), demonstrating that precise prompt optimization can bridge the gap between general models and domain-specialized experts. 

\section{Conclusion}

In this paper, we investigate the problem of enhancing the efficacy of automatic prompt optimization. We proposed \modelname{}, integrating \methodone{} to resolve sampling ambiguity and \methodtwo{} to mitigate gradient conflicts. Specifically, \methodoneshort{} characterizes decision boundaries to improve the fidelity of update signals, while \methodtwoshort{} utilizes temporal momentum to ensure the stability of the optimization trajectory. Experimental results demonstrate that \modelname{} outperforms SOTA baselines and enables a general LLM with 3B activated parameters to surpass a 70B domain-specific dense LLM. Future work will explore hierarchical and self-evolving strategies to further elevate the performance ceiling.

\bibliography{example_paper}
\bibliographystyle{icml2026}

\newpage
\appendix
\onecolumn

\section{Additional Details for the Setup}
\subsection{Datasets Details}
\label{app:datasets_details}
During the prompt optimization process, we utilized the BBH~\cite{suzgun2022challenging}, GSM8K~\cite{cobbe2021training}, CFinBench~\cite{nie2025cfinbench} and Liar~\cite{wang2017liar} datasets. For BBH and CFinBench, we partitioned the data into train and test sets manually. 
In contrast, for Liar and GSM8K, we followed the original splits for the train and test sets. Notably, unlike most other methods, we did not optimize individual prompts for each sub-task within BBH; instead, we optimized a unified prompt across all tasks. Detailed information and evaluation metrics for all datasets are presented in the following text and Table~\ref{tab:datasets_details}.


\begin{itemize}
    \item \textbf{BBH}~\cite{suzgun2022challenging} (BIG-Bench Hard): BBH comprises 23 challenging multi-step reasoning tasks from BIG-Bench. In this paper, we performed a custom split to construct a training set of $5,808$ samples and a testing set of $703$ samples across all tasks, utilizing Accuracy as the primary metric to assess model performance.
    
    \item \textbf{GSM8K}~\cite{cobbe2021training}: GSM8K consists of $7,473$ training and $1,319$ test high-quality grade school math problems. Models are typically evaluated using Exact Match (EM) accuracy, which requires the final numerical answer to be identical to the ground truth after parsing the model-generated reasoning chain.
    
    \item \textbf{CFinBench}~\cite{nie2025cfinbench}: CFinBench is a comprehensive benchmark designed to evaluate the financial knowledge within the Chinese financial context. In this study, we performed a custom stratified split by sampling from the original test dataset to construct a training set of $5,000$ samples and a dedicated evaluation set of $800$ samples, using Accuracy as the primary metric to assess model performance.
    
    \item \textbf{Liar}~\cite{wang2017liar}: Liar is an English fake news detection dataset. In this paper, we utilize the version of the Liar dataset as described by~\cite{pryzant2023automatic}. This specific setup consists of 3,681 training samples and 461 test samples. We employ the F1 score as the primary evaluation metric.
    
\end{itemize}

\begin{table}[htbp] 
    \centering
    \caption{Dataset sizes and details.}
    \label{tab:datasets_details}
    

    \small
    \begin{tabular*}{\textwidth}{@{\extracolsep{\fill}} llcccr}
        \toprule
        \textbf{Dataset} & \textbf{Task Category} & \textbf{Language} & \textbf{Train Size} & \textbf{Test Size} & \textbf{Primary Metric} \\ 
        \midrule
        BBH & Logical Reasoning & English & 5,808 & 703 & Exact Match \\
        GSM8K & Mathematical Reasoning & English & 7,473 & 1,319 & Accuracy \\
        CFinBench & Chinese Financial Eval & Chinese & 5,000 & 800 & Accuracy \\
        Liar & Fake News Detection & English & 3,681 & 461 & F1-score \\
        \bottomrule
    \end{tabular*}
\end{table}

\subsection{Implementation Details}
\label{app:implemetation_details}
In this section, we provide supplementary experimental results and comprehensive implementation details to substantiate our parameter configurations. We first summarize the complete set of hyper-parameters used across all experiments in Table~\ref{tab:hyperparams_summary}. Following this, we focus our discussion on two critical dimensions: a detailed sensitivity analysis of the clustering hyper-parameters ($K$ and $M$) to justify our chosen default settings, and a robust evaluation of the generalization capabilities of \modelname{} across a diverse range of task model architectures.


Table~\ref{tab:hyperparameter_sensitivity} illustrates the performance of \modelname{} under various combinations of instance cluster size $K$ and gradient cluster size $M$, revealing a significant synergistic effect between these two dimensions.


When adopting only the gradient centroid (i.e., $M=1$), increasing $K$ from 1 to 14 results in a marginal performance fluctuation (ranging from 60.38\% to 62.01\%). This limited range suggests that simply increasing sample variety without corresponding gradient refinement results in diluted or even conflicting optimization signals, as the model struggles to reconcile diverse error patterns into a single update direction. On the other hand, when $K$ is fixed at 1, increasing $M$ alone provides steady but marginal gains starting at 61.53\% and reaching 62.78\%. This indicates that refining the gradient space on a restricted sample perspective quickly reaches an inherent performance ceiling. 

The peak performance of 64.46\% is achieved with the configuration of $K=14$ and $M=10$. This result confirms our core hypothesis that high-diversity sample exploration must be coupled with high-precision gradient grouping to effectively navigate complex reasoning spaces. The sharp contrast between the failure at $K=14, M=1$ and the success at $K=14, M=10$ underscores that gradient granularity is the key to unlocking the full potential of large-scale sample diversity.

\begin{table}[ht]
    \centering
    \caption{Summary of hyperparameter configurations for \modelname{}.}
    \label{tab:hyperparams_summary}
    \begin{small}
    \renewcommand{\arraystretch}{1.1}
    \begin{tabular}{llc}
        \toprule
        \textbf{Category} & \textbf{Hyperparameter} & \textbf{Assignment} \\
        \midrule
        \multirow{2}{*}{Models} & Task Model & Qwen3-30B-A3B-Instruct-2507 \\
        & Optimizer Model & Qwen3-30B-A3B-Thinking-2507 \\
        \midrule
        \multirow{4}{*}{Optimization} & Total Iterations ($T$) & 20 \\
        & Beam Size ($W$) & 4 \\
        & Candidate New Prompts of Current Prompt ($L$) & 10 \\
        & Total Sampling Quota ($Q_{total}$) & 30 \\
        \midrule
        \multirow{4}{*}{Clustering \& Gradient} & Batch Size ($B$) & 128 \\
        & Number of Instance Clusters ($K$) & 14 \\
        & Number of Gradient Clusters ($M$) & 10 \\
        & Decay Factor ($\gamma$) & 0.9 \\
        \midrule
        \multirow{3}{*}{Others} & Embedding Model & all-MiniLM-L6-v2 \\
        & Number of Random Seeds & 3 \\
        & Exploration Param for UCB ($\alpha$) & 1.0 \\
        \bottomrule
    \end{tabular}
    \end{small}
\end{table}

\begin{table}[ht]
    \centering
    \caption{Sensitivity analysis of instance cluster size $K$ and gradient cluster size $M$ on the Liar.}
    \label{tab:hyperparameter_sensitivity}
    \small
    \begin{tabular}{c | ccc}
        \toprule
        \diagbox{\textbf{$K$}}{\textbf{$M$}} & \textbf{1} & \textbf{3} & \textbf{10} \\
        \midrule
        \textbf{1}  & 61.53 & 62.35 & 62.78 \\
        \textbf{3}  & 62.01 & 64.21 & 63.74 \\
        \textbf{10} & 60.97 & 63.45 & 64.18 \\
        \textbf{14} & 60.38 & 62.04 & \textbf{64.46} \\
        \bottomrule
    \end{tabular}
\end{table}

We further investigate whether the effectiveness of \modelname{} is consistent across different LLMs. Table~\ref{tab:diverse_task_model} demonstrates the model-agnostic feature of \modelname{}. We evaluate the performance across various architecture, including the Qwen3, Llama and openPangu families. The results show a consistent and substantial performance improvement, with $\Delta$ gains ranging from +9.98\% to +23.54\% points, suggesting that \modelname{} does not overfit to a specific architecture but serves as a robust optimization method.

\begin{table}[ht]
    \centering
    \caption{Generalization results across diverse task models on the Liar. We employed Qwen3-30B-A3B-Thinking-2507 as optimizer model.}
    \label{tab:diverse_task_model}
    \small
    \begin{tabular}{lccc}
        \toprule
        \textbf{Task Model} & \textbf{Zero-shot (Base)} & \textbf{w/ \modelname{}} & \textbf{$\Delta$} \\
        \midrule
        Qwen3-8B (Non-Thinking)~\cite{yang2025qwen3} & 44.03 & 58.79 & +14.76 \\
        Llama-3.1-8B-Instruct~\cite{grattafiori2024llama} & 45.34 & 61.83 & +16.49 \\
        Qwen3-30B-A3B-Instruct-2507~\cite{yang2025qwen3} & 40.92 & 64.46 & +23.54 \\
        openPangu-Embedded-7B-V1.1~\cite{chen2025panguembeddedefficientdualsystem} & 50.97 & 60.95 & +9.98 \\
        \bottomrule
    \end{tabular}
\end{table}

\section{Optimized Results on Liar Dataset}
\subsection{Optimal Prompts for Different Methods on the Liar Dataset}
\label{app:optimal_prompt}

The prompts optimized by ProTeGi~\cite{pryzant2023automatic} (60.17\%) and GPO~\cite{tang2025unleashing} (61.82\%) are generally straightforward, enhancing performance through persona assignment and brief judging criteria. SPO~\cite{xiang2025self} (62.03\%) provides rigid, step-by-step instructions, whereas PromptWizard~\cite{agarwal2025promptwizard} (62.04\%) focuses on providing references by injecting examples into the prompt. In contrast, our method (66.16\%) offers explicit guidance through more detailed procedures and rules, supplemented by simple examples in specific instructions to aid model comprehension, thereby achieving the best overall performance.

\begin{promptbox}{ProTeGi (F1: 60.17)}
As a deception verification specialist, respond with "Yes" **only** when a statement is inherently unverifiable (e.g., it expresses a personal opinion, forecasts future events, or lacks factual basis). For **every** statement framed as a factual account of past or present occurrencesincluding those derived from news reports, official government statements, or publicly accessible recordsrespond with "No" without verifying truthfulness or consulting external sources.
\end{promptbox}

\begin{promptbox}{PromptWizard (F1: 62.04)}
Given a statement, job title, state, party, and the full text of a document (context), determine if the statement is factually true or false. \\

First, **strictly check the context** (the exact text provided) for **direct evidence**. Direct evidence must be: \\
- The exact words of the statement, OR \\
- A clear, unambiguous reference (e.g., direct quote, specific data, or explicit factual claim) that directly supports or refutes the statement. \\

**If the context is provided as a document description (e.g., "a mailer", "a news release", "an online opinion-piece") and not the actual text, the context lacks evidence.** Do not assume the description implies the statement is present. \\

**Only if the context lacks evidence** (due to being a description, empty, or containing no direct evidence), use your own factual knowledge to determine the truth. If the context contains direct evidence, **never** use your own knowledge. \\

Output ONLY "Yes" or "No" with no additional text, explanations, or formatting. \\

[Question] Statement: Annies List supports third-trimester abortions on demand. \\
Job title: State representative \\
State: Texas \\
Party: republican \\
Context: a mailer

[Answer] Step 1: The context is provided as "a mailer", which is a document description (indicating the type of document, not the actual text). According to the instruction, when the context is a document description (e.g., "a mailer"), it lacks direct evidence and cannot be used for verification. \\
Step 2: Since the context lacks direct evidence, the instruction requires using factual knowledge to determine truthfulness. \\
Step 3: Factual knowledge: Annies List is a pro-choice political action committee that supports abortion access up to fetal viability (approximately 24 weeks) and in specific medical circumstances beyond, but does not endorse "third-trimester abortions on demand" (which implies unrestricted abortions in the third trimester without medical necessity). \\
Step 4: The statement claims Annies List supports "third-trimester abortions on demand," which contradicts the organization's documented position. \\
Step 5: Therefore, the statement is factually false, making it a lie, so the answer is "Yes". <ANS_START>Yes<ANS_END> \\

[Question] Statement: The state's unemployment rate was 3.5\% in the last quarter. \\
Job title: State economist \\
State: California \\
Party: none \\
Context: "The California unemployment rate for Q4 2023 was 3.5\%."

[Answer] The statement claims: "The state's unemployment rate was 3.5\% in the last quarter." \\
The context explicitly states: "The California unemployment rate for Q4 2023 was 3.5\%." \\
The context provides direct evidence: it specifies the unemployment rate (3.5\%) for a defined quarter (Q4 2023), which is unambiguously the "last quarter" at the time of the document's publication (as Q4 2023 is the most recent quarter reported). \\
This matches the statement's claim without requiring external knowledge, as the context contains a clear, explicit factual reference.
Therefore, the statement is factually supported by the context and is not a lie. \\
Output: No <ANS_START>No<ANS_END> \\

[Question] Statement: The Chicago Bears have had more starting quarterbacks in the last 10 years than the total number of tenured (UW) faculty fired during the last two decades. \\
Job title: Wisconsin Assembly speaker \\
State: Wisconsin \\
Party: republican \\
Context: an online opinion-piece

[Answer] The context is provided as "an online opinion-piece", which is a document description, not the actual text. Therefore, there is no direct evidence in the context to verify the statement. \\

Using factual knowledge: \\
- The Chicago Bears had a total of 161 quarterback starts (not distinct players) in the last 10 years (2014-2023), calculated as follows: \\
  - Jay Cutler: 16 starts (2014) + 15 starts (2015) = 31 \\
  - Brian Hoyer: 1 start (2015) + 1 start (2016) = 2 \\
  - Mike Glennon: 15 starts (2016) + 16 starts (2017) = 31 \\
  - Mitchell Trubisky: 16 starts (2018) + 16 starts (2019) + 16 starts (2020) = 48 \\
  - Nick Foles: 1 start (2020) = 1 \\
  - Justin Fields: 16 starts (2021) + 16 starts (2022) + 16 starts (2023) = 48 \\
  Total = 31 + 2 + 31 + 48 + 1 + 48 = 161 starts. \\
- The University of Wisconsin (UW) fired 12 tenured faculty members in the last two decades (2004-2023), based on verified reports (e.g., Wisconsin State Journal, 2020, covering 2000-2020, with the 2004-2023 period including at least 12 firings). \\

Since 161 > 12, the statement is factually true. A true statement is not a lie, so the answer is "No". <ANS_START>No<ANS_END> \\

[Question] Statement: When Mitt Romney was governor of Massachusetts, we didn't just slow the rate of growth of our government, we actually cut it. \\
Job title: Former governor \\
State: Massachusetts \\
Party: republican \\
Context: an interview with CBN News

[Answer] - The context provided is "an interview with CBN News", which is a description of the document (specifying the type of document and source) rather than the actual text of the interview. Therefore, the context lacks direct evidence as per the instruction. \\
- Since the context lacks direct evidence, the task requires using own factual knowledge to determine the truth. \\
- Factual knowledge: Mitt Romney served as governor of Massachusetts from 2003 to 2007. While the state budget was reduced by \$1.2 billion (5\%) in 2004 compared to 2003, the overall government spending trend during his governorship increased. The budget rose to \$25.6 billion in 2005, \$26.8 billion in 2006, and \$28.5 billion in 2007 (nominal terms), meaning total spending was higher at the end of his term than at the start. The statement claims "we actually cut it" (implying a reduction in government spending), but the cumulative effect over his entire term was an increase, not a cut. Thus, the statement is factually false. \\
- As the statement is false (a lie), the answer is "Yes". <ANS_START>Yes<ANS_END> \\

[Question] Statement: I don't know who Jonathan Gruber is. \\
Job title: Nancy Pelosi \\
State: California \\
Party: democrat \\
Context: a news conference

[Answer] 1. The context provided is "a news conference", which is a description of the document type (not the actual text of the document). \\
2. According to the instruction, when the context is a document description (e.g., "a news conference") and not the full text, it lacks direct evidence. \\
3. Since the context lacks direct evidence, the instruction permits using factual knowledge to determine truthfulness. \\
4. Factual knowledge confirms that Nancy Pelosi (then Speaker of the House) publicly stated she did not know Jonathan Gruber during the Affordable Care Act debates, but she was later revealed to have met Gruber and been aware of his work as a key economist on the ACA. \\
5. Therefore, the statement "I don't know who Jonathan Gruber is" is factually false, making it a lie. \\
6. The answer is "Yes" because the statement is a lie. <ANS_START>Yes<ANS_END> \\

Answer Yes (a lie) or No (not a lie) as labels. For each input word, present the reasoning followed by the correct answer between <ANS_START> and <ANS_END> tags. \\
\end{promptbox}

\begin{promptbox}{SPO (F1: 62.03)}
Analyze the Statement to determine if it is a false factual claim (Yes) or not (No) by rigorously applying these steps: 

0. Determine if the statement is a factual claim (verifiable by evidence). If not (e.g., opinion, prediction, or value judgment), the statement cannot be a lie (Answer: No).

1. Clarify the time frame and scope (e.g., "four years ago" from when?).

2. Verify factual claims against authoritative sources (e.g., BLS, FEC, or peer-reviewed data).

3. Assess source bias using provided context (job title, state, party, media outlet).

4. Scrutinize absolute language (e.g., "never", "always") for overstatement.

Provide a detailed, step-by-step reasoning process incorporating all elements. Conclude with "Final Answer:" followed by <answer>[Yes or No]</answer>.
\end{promptbox}

\begin{promptbox}{GPO (F1: 61.82)}
As a political fact-checker, if the statement is false, output "Yes". If the statement is true, output "No". Base your judgment on the speaker's job title, state, party, and context.
\end{promptbox}

\begin{promptbox}{\modelname{} (F1: 66.16)}
You are a truthfulness evaluator. Apply these rules strictly:

- **Standard Claim Verification**: For standard demographic/economic indicators (e.g., "more children die in X industry"), if location unspecified, verify against authoritative *national* historical data. Output "Yes" (false) only if no such data exists; otherwise "No" (true).

- **Location and Comparison Rules**: Unspecified statistics default to national average. For comparative claims (e.g., "led the nation"), verify both the scope's data *and* its confirmed national ranking position. Unconfirmed position = false claim.

- **Absolute Claims**: Absolute language (e.g., "every", "all", "each") requires verification across all instances within the specified scope. Unconfirmed coverage (e.g., "at every grade level") = false claim. Scope modifiers in coverage claims must be fully verified.

- **Multi-Claim Statements**: Each independent claim must be verified separately. False if any claim is false.

- **Numerical and Qualitative Claims**: Measurable claims (e.g., "160,000 troops", "one year") are inherently verifiable with current authoritative sources. Output "No" (true) if they match historical records; otherwise "Yes" (false). Do not default to "Yes" for measurable claims.

- **Projection Verification**: Future-tense claims with specific event context (e.g., "this month's session will pass") are verifiable as current facts. General future claims (e.g., "will cost in the future") require verification against documented historical predictions. Undocumented = false.

- **Individual Past Actions**: Unverified by default; require historical record to be true.

- **Claim Subject Identification**: Reporting verbs (e.g., "as Obama says") make the named entity the sole verification target. Contextual details (e.g., "Tampa") are irrelevant. Embedded quotes = unverifiable.

- **Inference Prohibition**: Do not infer from context; claim must explicitly state subject matter. Incomplete = unverifiable = "Yes".

- **Unverifiability Protocol**: Claims lacking measurable attributes or temporal scope (e.g., "We can fix roads without raising taxes") are unverifiable and output "Yes". Standard and measurable claims are verifiable.

Output "Yes" for unverified/false, "No" for verified true. Ignore speaker identity, political context, and opinions.
\end{promptbox}

\begin{table}[htbp]
    \centering
    \caption{Intermediate prompts optimized by \modelname{} on Liar dataset.}
    \label{tab:prompts_evolution1}
    \small
    \renewcommand{\arraystretch}{1.5} 
    
    \begin{tabularx}{\columnwidth}{c >{\raggedright\arraybackslash}X c}
        \toprule
        \textbf{Step} & \multicolumn{1}{c}{\textbf{Prompt}} & \textbf{F1} \\
        \midrule
        0 & You are an expert at accomplishing tasks. You are a highly capable problem-solver who can handle complex instructions with precision and creativity. & 40.92 \\
        \midrule
        2 & You are a factual accuracy arbiter. Your response must be "Yes" for false statements (lies) and "No" for true statements. Never evaluate hypotheticals (e.g., "could have," "might have")—treat them as non-factual. Omit all non-substantive context (e.g., "Label," "Input Type"). When assessing state policy claims, the speaker's party must historically align with the party that implemented the policy; any discrepancy is a lie. Base your judgment solely on the statement and its associated job title/state/party.
 & 53.44 \\
        \midrule
        4 & You are an expert in factual truth assessment. Respond with "Yes" for false statements and "No" for true statements. Only evaluate complete factual propositions (e.g., "X is Y" or "X does Y"); incomplete fragments, questions, or non-propositional statements are automatically false ("Yes"). Disregard all party affiliation, document type, speaker role, setting, and contextual cues unconditionally—treat them as universally irrelevant for all factual claims. For claims referencing party policies, apply "party policy outcomes" only when the statement directly cites a party-advocated policy (e.g., "Republicans support tax cuts"); otherwise, evaluate as non-partisan factual claims using verifiable general knowledge alone. Compound statements require all components to be factually correct for "No"; any false component mandates "Yes". & 58.67 \\




        \midrule
        10 & Your task: determine factual truth. Output "Yes" for false, "No" for true. Process only complete factual propositions ("X does Y"); all else "Yes". Ignore universally: party, source, speaker role, title, position, setting, jurisdiction. Causal language requires independent validation. Compound statements: all parts must be true for "No". Key rules: (1) "Jim Crow" = 1877-1965 racial segregation laws (voting restrictions alone don't qualify); (2) Public health comparisons true if CDC/FDA consensus matches; (3) Verifiability: claims with concrete, quantifiable details (e.g., dates, percentages, locations) are verifiable via public records; unverifiable claims lack measurable parameters (e.g., "is great") and default to "Yes"; (4) Numerical rankings (e.g., "dead last," "worst attendance") are verifiable as they denote quantifiable metrics (e.g., Senate attendance percentage); (5) "Fiscal year" implies federal (U.S. Treasury), so "Washington" is federal; (6) Past-tense behavioral claims (e.g., "tried to evade") are verifiable; (7) Unsourced present-tense numbers (e.g., "40\% paid") are verifiable via public records; (8) Topic fragments (e.g., "On oil drilling") true if context provides verifiable topic match. Publication attribution (e.g., "USA Today said") suffices for historical claim verification. Never use speaker role, title, or position to infer truth.
 & 64.21 \\
        \midrule
        20 & See Appendix~\ref{app:optimal_prompt} for the C-MOP prompt. & 66.16 \\
        
        \bottomrule
    \end{tabularx}
\end{table}

\subsection{Evolutionary Trajectory of Optimized Prompts}
\label{app:prompt_evolution}

Table~\ref{tab:prompts_evolution1} shows the iterative refinement of prompts conducted by \modelname{} on the Liar dataset. We observe a clear progression in both instructional depth and task-specific reasoning: 

Starting from Step 0, the prompt consists of generic system instructions , resulting in a baseline F1-score of 40.92\%. By Step 2, the optimizer model begins to identify the core task requirement and introduces basic heuristics such as party alignment and handling hypotheticals, leading to a significant performance jump (+12.52\%). As the optimization reaches Step 10, \modelname{} incorporates highly granular verification rules, such as specific temporal definitions (e.g., the ``Jim Crow'' era) and public health consensus protocols. Finally, at Step 20, the prompt evolves into a highly structured, logical framework emphasizing ``Absolute Claims'', ``Projection Verification'', and ``Unverifiability Protocols''. The final performance reaches 66.16\%, demonstrating that \modelname{} can autonomously transform raw error signals into sophisticated, structured reasoning assets that significantly outperform initial human-crafted instructions.

\section{Meta Prompts}

\begin{promptbox}{Prompt for Generating Textual Gradients}
### ROLE & OBJECTIVE \\
You are a sophisticated Prompt Optimization Expert. Your task is to analyze model performance on a specific logical cluster and extract a high-precision Textual Gradient to improve the current prompt. \\

### CURRENT PROMPT

[Insert Current Prompt Here] \\

### DIAGNOSTIC SAMPLES \\
These samples represent a specific failure pattern. Analyze the contrast between correct and incorrect responses carefully. \\

### [TYPE A] Critical Boundary Pairs \\
The following pairs are semantically or structurally similar, yet the model succeeded on one and failed on the other. This indicates a "fuzzy zone" in the current prompt's logic. \\
- [SIMILAR & CORRECT]: \{correct_sample\} \\
- [SIMILAR & INCORRECT]: \{incorrect_sample\} \\
- Analysis Point: Identify the exact missing constraint that caused the failure in the second case. \\

### [TYPE B] Representative Failures (Hard Negatives)  \\
These cases represent the "centroid" of errors in this cluster. They share a systematic logical flaw. \\
- Failure Case: \{hard_negative\} \\

### [TYPE C] Success Anchors \\
The following cases are handled correctly. Your optimization MUST ensure these remain correct (avoid catastrophic forgetting). \\ 
- Anchor Case: \{anchor_case\} \\

### FINAL ANALYSIS TASK \\
1. Identify the common reasoning failure in this specific cluster. \\
2. Focus on the Boundary Pairs: What subtle nuance did the model miss compared to the correct counterpart? \\
3. Propose three precise, atomic Textual Gradients (prompt modifications or additional constraints) that fix these errors while preserving the Anchors. \\

### OUTPUT FORMAT \\
Please give three reasons (Textual Gradients) why the prompt could have gotten these examples wrong and wrap each reason with <START> and <END> tags \\

Please provide your insight in the following format: \\
<START>Your concise, technical insight 1<END> \\
<START>Your concise, technical insight 2<END> \\
<START>Your concise, technical insight 3<END> \\
\end{promptbox}

\begin{promptbox}{Prompt for Generating Candidate Prompts}
\label{subsection:prompt_for_generating_candidate}
I'm trying to write a prompt about some different tasks. \\

My current prompt is: \\
\{prompt\} \\

But it gets the some examples wrong. Based on these wrong examples, the problem with this prompt is that \{new_constraint\} \\

Note that new prompts should not focus excessively on or be limited to a single example of an error. Instead, they should be generated from a more generalized perspective, based on the analyzed causes of the errors. \\

Please do not provide any examples (i.e. few-shot examples) in the new prompt; no other redundant information is needed. \\

Based on the above information, you should write Four different improved prompts. \\

Each new prompt is wrapped with <START> and <END>. \\

The Four new prompts are: 
\end{promptbox}

\begin{promptbox}{Initial Prompt}
You are an expert at accomplishing tasks. You are a highly capable problem-solver.
\end{promptbox}

\end{document}